\documentclass{article}

\usepackage{arxiv}

\usepackage[utf8]{inputenc} 
\usepackage[T1]{fontenc}    
\usepackage{hyperref}       
\usepackage{url}   
\usepackage{float}
\usepackage{booktabs}       
\usepackage{amsfonts}       
\usepackage{nicefrac}       
\usepackage{microtype}      
\usepackage{lipsum}
\usepackage{graphicx}
\usepackage{multirow}
\usepackage{tablefootnote}
\usepackage{subcaption}
\usepackage[skip=0pt]{caption}
\graphicspath{ {./figures/} }

\title{Leveraging Segment Anything Model in Identifying Buildings within Refugee Camps (SAM4Refugee) from Satellite Imagery for Humanitarian Operations}

\author{
 Yunya Gao \\
  The Christian Doppler Laboratory for Geospatial and EO-based Humanitarian Technologies\\
  Department of Geoinformatics–ZGIS\\
  Paris Lodron Universität Salzburg\\
  5020 Salzburg, Austria\\
  \texttt{yunya.gao@plus.ac.at} \\
}

\begin{document}
\maketitle
\begin{abstract}
Updated building footprints with refugee camps from high-resolution satellite imagery can support related humanitarian operations. This study explores using the "Segment Anything Model" (SAM) and one of its branches, SAM-Adapter, for semantic segmentation tasks in building extraction from satellite imagery. SAM-Adapter is a lightweight adaptation of the SAM and emerges as a powerful tool for this extraction task in diverse refugee camps. Our research proves that SAM-Adapter excels in scenarios where data availability is limited compared to other classic (e.g., U-Net) or advanced semantic segmentation models (e.g., Transformer). Furthermore, the impact of upscaling techniques on model performance is highlighted, with methods like super-resolution (SR) models proving invaluable for improving model performance. Moreover, the study unveils intriguing phenomena, including the rapid convergence of the model in the first training epoch when using upscaled image data for training, suggesting opportunities for future research. The codes covering each step from data preparation, model training, model inferencing, and the generation of Shapefiles for predicted masks are available on a \href{https://github.com/YunyaGaoTree/SAM-Adapter-For-Refugee-Dwelling-Extraction}{GitHub repository} to benefit the extended scientific community and humanitarian operations. 
\end{abstract}

\keywords{remote sensing \and deep learning \and segment anything \and refugees}

\section{Introduction}
The United Nations Statistical Commission endorsed various indicators for refugees as part of the Sustainable Development Goals (SDGs), adhering to the "Leave no one behind" commitment ~\cite{UNHCR2020Sustainable}. SDGs 2, 3, 6, and 7 promote providing essential living resources such as clean water, healthy food, healthcare services, contemporary energy, and hygiene to refugees and the communities that host them. It's crucial to understand the population of refugees in need before distributing these resources. Given the challenges in gathering this information directly on the ground, utilizing updated satellite imagery of refugee camps can be advantageous for estimating purposes ~\cite{Checchi2013Validity, Spröhnle2014Earth, Füreder2014Object-based}.

Recently, deep learning techniques have shown significant promise in extracting building footprints within refugee camps from satellite images ~\cite{Lang2020Earth, Tiede2017Stratified}. However, most studies rely on strong supervision, which requires a substantial number of high-quality annotations in advance. For example, ~\cite{Gao2022Comparing} compares multiple classical semantic segmentation networks for the extraction task. The findings confirm that many networks can produce encouraging results and exhibit comparable performance when subjected to strong supervision for this extraction task. Given the high cost of acquiring pixel-precise annotations, there's an urgent need to develop both label-efficient and effective approaches, especially for humanitarian operations aimed at assisting refugees.

In recent times, there has been a rising interest in foundation models. This surge in attention can be linked to their extensive pre-training on large datasets and their exceptional capability to adapt to a wide range of downstream tasks ~\cite{Ji2023Segment}. Recently, Meta AI Research unveiled a promptable model called the "Segment Anything Model" (SAM) ~\cite{Kirillov2023Segment}. SAM is trained on the vast SA-1B dataset, encompassing over 11 million images and 1 billion masks. It has demonstrated impressive zero-shot transfer capabilities without the need for fine-tuning ~\cite{Zhang2023Survey}.  

Although SAM excels in generalizing typical scenes, it struggles to detect smaller, irregular structures such as buildings or roads on satellite imagery ~\cite{Ji2023Segment}. In the building extraction task, SAM does not outperform task-specific models such as U-Net ~\cite{Ren2023Segment}. Due to these limitations, customizing SAM for building extraction tasks can be of great importance. Therefore, this study utilizes SAM-Adapter for the building extraction for refugee camps, framed as a semantic segmentation task with a limited number of pixelwise labeled samples in terms of spatial extent. 

SAM-Adapter is lightweight, and can easily adapt the large pre-trained foundation model of SAM to specific downstream tasks by adapters ~\cite{Wu2023Medical}. We evaluated the performance of SAM-Adapter against other classical and cutting-edge semantic segmentation models, using two different data sizes sourced from five different refugee camps. Satellite imagery from the chosen study locations comes from various sensors and displays various spatial resolutions. The results from this study reveal that SAM-Adapter greatly surpasses other semantic segmentation models especially when using limited sample data across all study sites. 

Furthermore, multiple studies have determined that zoom levels can profoundly impact the efficacy of deep learning methods in tasks related to building extraction ~\cite{Touzani2021Open, Hoffmann2021Zooming}. In addition, combining SAM with the SR (super-resolution) model increases its proficiency for low-resolution images ~\cite{Osco2023Segment}. Inspired by this finding, we evaluated the performance of SAM-Adapter using datasets with and without upscaling by SR models. 

The main contributions are shown below:

    1) This study is pioneering in adapting SAM specifically for refugee dwelling extraction tasks using limited pixelwise sample data in terms of spatial extent. The results confirm that SAM-Adapter surpasses other traditional and cutting-edge semantic segmentation models.
    
    2) Given the requisite input size of SAM, enhancing the zoom level by upscaling image data yields better results than other data-augmentation techniques, such as rotation. The SAM-Adapter, when trained with upscaled images, delivers the best outcomes right from the first epoch. This discovery requires further exploration in future studies.
    
    3)	The entire workflow, covering every step from data preparation (including the upscaling of satellite imagery in GeoTIFF format) to model training, evaluation, and the generation of predicted masks in the format of polygonal Shapefiles containing geospatial information, is accessible within the \href{https://github.com/YunyaGaoTree/SAM-Adapter-For-Refugee-Dwelling-Extraction}{GitHub repository}.

\section{Related work}
\label{sec:headings}
To date, studies exploring the use of SAM to extract building footprints from satellite imagery remain sparse. ~\cite{Ren2023Segment} pioneered the examination of the ability of the SAM to extract buildings and different objects using diverse prompts (e.g., center points, random points, grid points, and bounding boxes). Yet, the findings indicate that SAM does not surpass task-specific models like U-Net based on these prompts. Furthermore, the accuracy for buildings is considerably lower than for other objects such as solar panels or clouds.

~\cite{zhang2023text2seg} introduce a pipeline named Text2Seg to tackle semantic segmentation tasks in satellite imagery. Text2Seg integrates SAM with other visual learning models such as Grounding DINO ~\cite{Liu2023Grounding}. Although the prompt tuning process demands minimal effort, making it operationally favorable, the predicted masks from Text2Seg often result in a lot of errors in building extraction.

~\cite{chen2024rsprompter} introduce RSPrompter, an automated instance segmentation method tailored for remote sensing images. The RSPrompter is built on the SAM foundation model and integrates semantic category data. Drawing inspiration from prompt learning, they propose a prompt generator designed to learn to create suitable prompts for SAM input. This approach aims to diminish semantic disparities and avoid the overfitting of the head. The generator is inherently linked to category information, yielding semantic instance segmentation outputs. Furthermore, the results indicate that fine-tuning the SAM decoder with minimal data might not always be the best course of action.

~\cite{Osco2023Segment} develop a new automated method that merges a general text prompt-based example with one-shot training using PerSAM ~\cite{Zhang2023Personalize}. The building class accuracy improves significantly when compared to SAM which is only adjusted using bounding box prompts. This research also notes that SAM's performance changes on the basis of the input imagery's spatial resolution, with more errors at lower resolutions. Hence, combining SAM with SR models may boost its ability to work with low-resolution images.

In summary, using prompts like points or bounding boxes can enhance the effectiveness of SAM in extracting buildings from satellite images. Yet, they don't match the performance of task-specific models like U-Net. While one-shot training can boost SAM's building extraction abilities, merely adjusting the SAM decoder is not always recommended. Combining SAM with the SR method may improve its proficiency with low-resolution images. 

Based on the above findings, we chose to implement domain-specific adapters for SAM, termed SAM-Adapter ~\cite{chen2023sam}, to tackle the refugee dwelling extraction challenge. In the SAM-Adapter, trainable adapters are attached between the transformer layers of the SAM image encoder. As a result, fine-tuning is applied not only to the mask decoder but also to the adapters. The foundation model remains frozen, while the adapters undergo fine-tuning using limited pixelwise samples. Furthermore, we assess the impact of various upscaling methods, nearest interpolation, bilinear interpolation, and SR models, on SAM-Adapter's efficacy. This examination aims to determine whether the use of traditional upscaling techniques or SR models can elevate the performance of the SAM adapter as suggested by ~\cite{Osco2023Segment}.

\section{Methodology}
\subsection{Study sites}
To assess the generality and transferability of SAM-Adapter in the task of refugee dwelling extraction, we select five distinct refugee camps, each showcasing different types of building roof ontology. The chosen sites are: 1) Kututpalong refugee camp, 2) Nduta refugee camp, 3) Dagahaley refugee camp, 4) Djibo refugee camp, and 5) Minawao refugee camp.

Kutupalong refugee camp, located in the Cox's Bazar region of Bangladesh, serves as a major refuge for Rohingya refugees who have been fleeing persecution from Myanmar. Its significance has grown over time, especially in 2017 due to the heightened influx of refugees. As of July 2023, with more than 931,000 refugees, Kutupalong has become the world's largest refugee camp ~\cite{UNHCR2023Inside}. 

The Nduta refugee camp is located in the North-West area of Tanzania ~\cite{UNHCR2018North-West}. It was set up on October 4, 2015, primarily as an emergency measure to house 40,000 Burundian refugees. These refugees were being moved from the overcrowded Nyarugusu refugee camp.

Dagahaley is one of the camps in the Dadaab refugee complex located in Kenya. The camp was established in March 1992. For its first 14 years of existence, the Dagahaley camp maintained a population of about 30,000 residents ~\cite{UNHCR2015Dagahaley}.

The northern town of Djibo in Burkina Faso has been under blockade by non-state armed groups for over a year, severely limiting access to food and aid for its residents. In the nearby Sahel region, Burkina Faso hosts a refugee camp with around 6,500 refugees, many of whom are from Mali ~\cite{UNHCR2020Burkina}.

Minawao refugee camp is situated in the Far North region of Cameroon. As of October 2021, it is recognized as one of Cameroon's largest refugee shelters, housing more than 68,000 Nigerian refugees ~\cite{UNHCR2022Annual}.

\subsection{Data processing}
\begin{figure}
    \centering
    \includegraphics[width=1\linewidth]{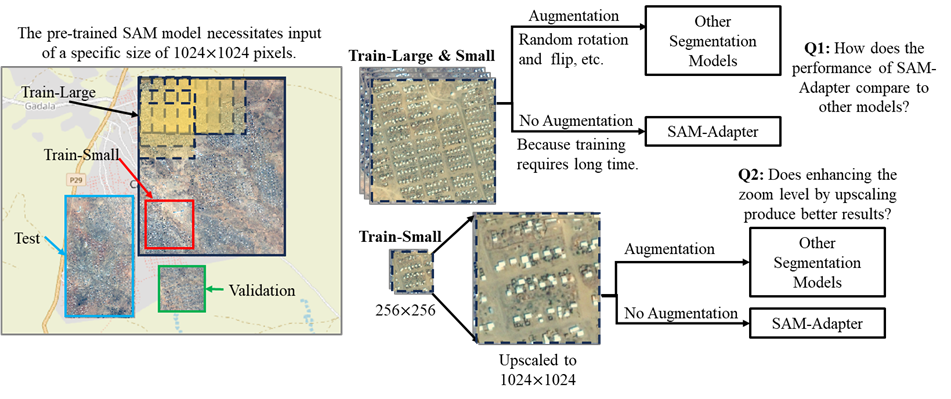}
    \caption{A general workflow of data sampling and preprocessing.}
    \label{fig:fig1}
\end{figure}

The general data processing workflow is illustrated in Figure \ref{fig:fig1}. Satellite image data for the five study sites originate from various sensors, with differing spatial resolutions and capture dates, as outlined in Table \ref{tab:table1}. All image data, except for Kutupalong, have been subjected to pansharpening through the panchromatic band. We assess the transferability of SAM-Adapter across a range of scenarios, utilizing two dataset sizes (Train Large and Train Small). Due to differences in spatial resolutions, the image data from the Kutupalong refugee camp (0.1 m) can generate more data patches when maintaining a spatial extent similar to other sites. Consequently, we chose a broader area for the Nduta refugee camp to match, as closely as possible, the pixel-level extent of the Kutupalong refugee camp. Given that the building roof ontology of the Nduta refugee camp is notably simpler than that of Kutupalong, our objective is to determine which variable - the number of training patches, the spatial resolution and other factors - has a more significant impact on the extraction process.

\begin{table}[ht]
    \centering
    \scriptsize
    \caption{Detailed information of data for each refugee camp.}
    \label{tab:table1}
    
    \begin{tabular}{ccccccc}
        \toprule
        \textbf{Refugee camp} & \textbf{Retrieved date} & \textbf{Sensor} & \textbf{Resolution (m)} & \textbf{Data type} & \textbf{Extent/pixel} & \textbf{Nr. of patches} \\ \hline  
         
        Kutupalong & 13/02/2018 & Drone & 0.1 & Train\_Large & 13283, 12489 & 1848 \\
        & & & & Train\_Small & 6288, 5346 & 420 \\
        & & & & Validation & 4344, 4079 & 226 \\
        & & & & Test & 9494, 4020 &  \\ \hline
        Nduta & 13/08/2017 & WorldView-3 & 0.3 & Train\_Large & 13947, 7624 & 1176 \\
        & & & & Train\_Small & 4101, 4188 & 224 \\
        & & & & Validation & 4094, 3177 & 63 \\
        & & & & Test & 6996, 6145 & \\ \hline
        Dagaheley & 08/04/2017 & WorldView-3 & 0.3 & Train\_Large & 5399, 5074 & 350 \\
        & & & & Train\_Small & 1914, 1049 & 56$^a$ /350$^b$ \\
        & & & & Validation & 4383, 1973 & 7 \\
        & & & & Test & 2127, 1998 & \\ \hline
        Djibo & 12/12/2019 & Pleiades-1A & 0.5 & Train\_Large & 1406, 599 & 280 \\
        & & & & Train\_Small & 2204, 2004 & 56/280 \\
        & & & & Validation & 2053, 1956 & 7 \\
        & & & & Test & 2000, 1991 &  \\ \hline
        Minawao & 12/02/2017 & WorldView-2 & 0.5 & Train\_Large & 3882, 4265 & 224 \\
        & & & & Train\_Small & 3621, 1458 & 56/224 \\
        & & & & Validation & 1888, 1773 & 7 \\
        & & & & Test & 1817, 3165 &  \\ \hline
        
        \bottomrule
    \end{tabular}
    \captionsetup{justification=raggedright, singlelinecheck=false, font=scriptsize}
    \caption*{$^a$ The number of patches without upscaling. \\
              $^b$ The number of augmented patches without upscaling.}

\end{table}

It's crucial to acknowledge that there are multiple methods for sampling image data. For instance, instead of maintaining a similar spatial extent for each study site, it may be worthwhile to keep the number of generated patches consistent. It may be valuable to explore different sampling strategies such as different proportions of data. However, in this initial research, we choose this straightforward data collection approach.

Each patch is intentionally overlapped with its neighboring patches. The overlap is instrumental in ensuring that no vital features are overlooked, particularly in regions with densely clustered buildings or intricate structures ~\cite{Milosavljević2020Automated}. During the inference process, overlapping patches are also applied for the testing data. This step is taken to prevent the buildings located at the edges of the individual patches from being separated or divided, which could result in incomplete or inaccurate extraction ~\cite{chen2023sam}.

The created patch data of “Train-Large” are directly fed to the SAM-Adapter. However, they are augmented by rotation, flipping, and random brightness contrasting before being fed to other semantic segmentation models. This is due to the relatively slow training speed of SAM-Adapter, even though the foundation model has been frozen. To provide an estimate, the training speed for one epoch of the SAM-Adapter is approximately ten times slower compared to other semantic segmentation models utilized in this study. However, to evaluate the impact of data augmentation on SAM-Adapter, we augment the created Train Small dataset by rotation and flipping. From the results based on these datasets, we aim to solve the following research question:

\textit{(1)	How does the performance of SAM-Adapter differ from other semantic segmentation models with two different dataset sizes?}

The pretrained SAM model requires a fixed input size of patches (1024×1024 pixels), which may be relatively large for refugee dwellings in lower spatial resolution images. Therefore we have created two types of Train Small datasets for Dagahaley, Djibo, and Minawao refugee camps. Because these three sites contain much less training data compared to the other two sites. The first dataset consists of image patches that are 1024×1024 pixels in size, which is the same for Train Large datasets. The second dataset starts with image patches that are 256×256 pixels, and then are upscaled to 1024×1024 pixels using nearest interpolation, bilinear interpolation, or an SR model.

Reducing image patch sizes can result in the generation of a greater quantity of patches. To determine if SAM-Adapter's performance alteration is impacted by upscaling or increased data volume, we decrease the overlapping steps for the second type of datasets to match the number of patches created in the Train Large datasets. Additionally, we maintain the patch count for augmented Train Small datasets at a level identical to that of the second type of datasets. From these experiments, we try to answer the research question below:

\textit{(2) Does enhancing the zoom level by upscaling produce better results for small datasets?}

\subsection{Increasing zoom level by upscaling}
Enhanced deep super-resolution network (EDSR) ~\cite{Lim2017Enhanced}, is selected through the OpenCV Python API ~\cite{Agarwal2021Super}. EDSR utilizes a multiscale deep-SR system to reconstruct higher-resolution images. The upscaling factor is set to 4 as provided in OpenCV. It has been chosen due to its open-source nature and efficiency. During preliminary experiments, we also tried the residual dense network ~\cite{Zhang2018Residual} and the cascading residual network ~\cite{Ahn2018Fast} as SR models for upscaling. The preliminary findings show that these SR models have a similar impact on SAM-Adapter's performance. As a result, we only retain the results from the EDSR. 

It should be noted that these SR models were originally designed for natural scene images, which limits their effectiveness for satellite imagery. To the best of our knowledge, current open-source SR models for satellite imagery do not support increasing spatial resolution greater than 0.5 m. For example, SR4RS trained by 250 different Spot-6 and Spot-7 scenes with a spatial resolution of 2.5 m is beneficial to enhance Sentinel-2 imagery ~\cite{cresson2022sr4rs}, which is far from our need. Therefore, they are not applied in this study.

\begin{figure}[h]
    \centering
    \includegraphics[width=0.8\linewidth]{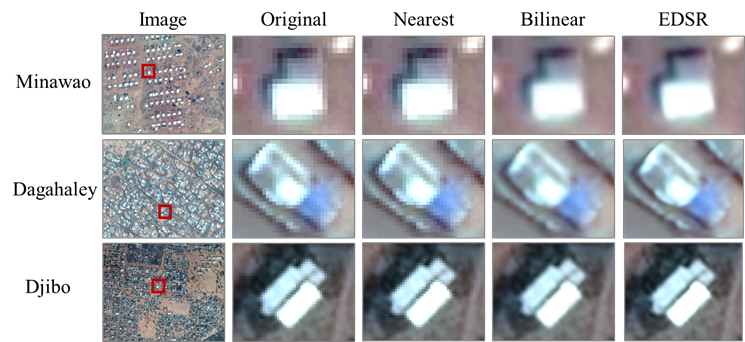}
    \caption{Visual Comparison of Upscaling Methods: Bilinear and Nearest Interpolation and EDSR for Minawao, Dagahaley and Djibo refugee camps.}
    \label{fig:fig2}
\end{figure}

Figure\ref{fig:fig2} shows the effects of the three upscaling methods. EDSR-enhanced images display sharper boundaries compared to those upscaled by bilinear interpolation. Images upscaled using nearest interpolation look much like the original and don't enhance the dwelling boundaries.

\subsection{SAM Adapter}
\begin{figure}[h]
    \centering
    \includegraphics[width=1\linewidth]{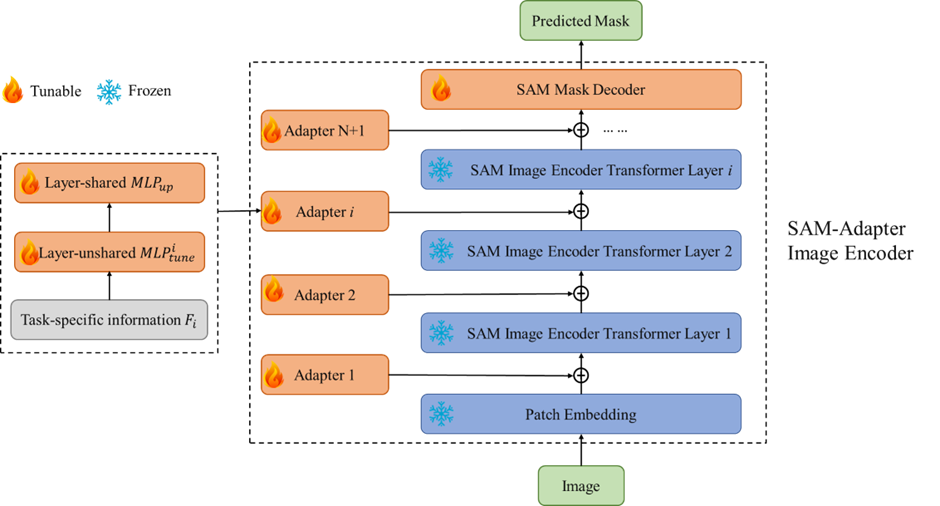}
    \caption{Brief architecture of SAM-Adapter.}
    \label{fig:fig3}
\end{figure}

SAM-Adapter leverages the SAM image encoder as the backbone of a segmentation network ~\cite{chen2023sam}. SAM's image encoder is based on a ViT-H/16 model, featuring a 14×14 windowed attention mechanism and four global attention blocks spaced equally ~\cite{Kirillov2023Segment}. The pretrained SAM image encoder weights remain frozen. SAM's mask decoder includes a modified transformer decoder block and a dynamic mask prediction head. Weight initialization for the mask decoder utilizes the pretrained SAM's weights. The decoder is fine-tuned during training. The brief SAM-Adapter's architecture is presented in Figure \ref{fig:fig3}.

Task-specific knowledge is incorporated into the network through adapters. A tunable adapter is added to the output of each transformer layer in the SAM image encoder ~\cite{chen2023sam}. Employing suitable prompts to introduce task-specific knowledge can significantly improve the model's ability to generalize to downstream tasks for foundation models, particularly when labeled data is limited ~\cite{Liu2023Explicit}. 

An adapter consists of two multilayer perceptron (MLP) and an activate function within two MLP ~\cite{Liu2023Explicit}. MLP is a type of modern feedforward artificial neural network ~\cite{Ramchoun2016Multilayer}. It is composed of fully connected neurons equipped with a non-linear activation function. One of its significant characteristics is the ability to distinguish between data that are not linearly separable. In essence, the MLP is designed to learn the relationship between linear and nonlinear data. Specifically, an adapter takes the information F and obtains the prompt P based on the following equation:

\begin{equation}
P_i=MLP_{up}(GELU(MLP_{tune}^i (F_{i})))
\end{equation}

where MLP are linear layers employed to generate task-specific prompts for each Adapter. It's a shared up-projection layer utilized by all Adapters to modify the dimensions of transformer features. P refers to the output prompt linked with each SAM model transformer layer. GELU represents the Gaussian Error Linear Unit activation function ~\cite{Hendrycks2016Gaussian}. Information F assumes different forms, depending on the task, and is adaptable in design. For instance, it can be derived from task-specific dataset samples, taking on forms such as texture or frequency information, or through manual rule creation.

To assess SAM-Adapter's performance, we conduct an ablation study by choosing various classical and advanced semantic segmentation models for comparison. These models include the Feature Pyramid Network (FPN) ~\cite{Lin2017Feature} with 1) Mix Transformer-B0 (MiT) from SegFormer ~\cite{xie2021segformer}, 2) MobileNet-v3-Large (MobileV3L) ~\cite{Howard2019Searching}, 3) ResNet34 ~\cite{he2016deep} as their backbones, and Unet ~\cite{ronneberger2015u} with 4) MobileNet-v3, 5) ResNet34, and 6) ResNet101 ~\cite{he2016deep} as their backbones. 

\subsection{Accuracy metrics}
We evaluate the performance of the proposed approach with Precision, Recall, F1-score, and IoU of refugee dwellings. The calculation of the four metrics can be found below, where TP, FP, and FN refer to the number of True Positive, False Positive, and False Negative pixels for the semantic class.

\begin{equation}
Precision =TP/(TP+FP)
\end{equation}
\begin{equation}
Recall=TP/(TP+FN)
\end{equation}
\begin{equation}
F1 =2*(Precision*Recall)/(Precision+Recall)
\end{equation}
\begin{equation}
IoU=TP/(TP+FP+FN)
\end{equation}

\subsection{Implementation details}
For a comparative analysis of the impact of fine-tuning, we present results from SAM without any fine-tuning. These predicted masks are derived using the Segment Geospatial Python API ~\cite{wu2023samgeo}. 

Most hyperparameter settings of SAM-Adapter follow the default setting in ~\cite{chen2023sam}. We change the batch size from 2 to 1 due to limited computation resources. “sam vit h 4b8939.pth” pretrained model is selected due to its better performance in building extraction tasks ~\cite{chen2024rsprompter}. AdamW is chosen as an optimizer with an initial learning rate of 2×10-4 and a minimal learning rate of 10-7. The total training epochs for SAM-Adapter is 15. An analysis has been done to analyze the influence of training epochs.

For other semantic segmentation models, RAdam ~\cite{Howard2019Searching} is selected as the optimizer with an initial learning rate of 10-3. For the learning rate scheduler, ReduceLROnPlateau is chosen for its ability to prevent model stagnation and dynamically adjust the learning rate. The patience is set to 5, and a learning rate reduction factor of 0.2 is applied. Batchsize is 8. The total number of training epochs is 50. The model with optimal performance on validation data is used for inference and evaluation of test data.

All of the experiments were conducted on PyTorch 2.0.1 and Python 3.9 environment. The networks undergo training and testing on a machine equipped with a NVIDIA RTX3090 GPU.

\section{Results and Discussion}
\subsection{Comparison SAM-Adapter with other models}
Table\ref{tab:table2} showcases the accuracy assessment results for the Kutupalong and Nduta refugee camps. Meanwhile, Table \ref{tab:table3} presents the results for the Dagahaley, Djibo, and Minawao refugee camps. The top results for each site and data size are marked in red in bold. The best results from models excluding SAM-Adapter are highlighted in blue bold. Figures \ref{fig:fig4} through 8 display visual representations of predicted masks for all study sites. In the Appendix, several figures display larger satellite images and predicted masks produced by SAM-Adapter and FPN-MiT when trained on small datasets without any upscaling. The supplementary materials are provided to offer a broader perspective on the different performance between SAM-Adapter and FPN-MiT in operations. 

First of all, we can observe that SAM, without any fine-tuning (noFT), falls short in performance compared to other models across all study sites. This aligns with findings from previous research indicating SAM's challenges in identifying buildings from satellite imagery ~\cite{Ren2023Segment}. We then utilize the SR model (EDSR) on the test data from the Dagahaley, Djibo, and Minawao refugee camps. Subsequently, we assess SAM's predictions (No-FT) on the upscaled test data. The findings suggest that, while the upscaled test data produce slightly improved results, the quality remains unsatisfactory. 
In the visual analysis, SAM (noFT) reveals signs of over-segmentation and an inability to adjust to dataset nuances, evident from scenarios like Kutupalong, where it segments roads and other non-building objects. Similarly, in Djibo, the model’s coarse segmentation of entire residential units as single objects points to challenges in recognizing building boundaries. In Minawao, the blanket segmentation of the entire input patch underscores an extreme oversegmentation. These flaws may arise from issues in imagery resolution or generic training without dataset-specific fine-tuning.

In terms of model performance on different data sizes, large datasets consistently yield better performance for all models in all refugee camps. Models like FPN-ResNet34, Unet-ResNet101, and Unet-MobileV3L show a substantial decrease in IoU when transitioning from large to small datasets. For instance, in the Dagahaley camp, the IoU for FPN-ResNet34 drops from 0.351 to 0.138, highlighting the model's dependency on data size. In contrast, the SAM adapter demonstrates relatively stable performance across data sizes. This suggests a potential robustness of the SAM-Adapter in scenarios with limited data. Take the same example of the Dagahaley camp, the IoU value changes from 0.619 (large dataset) to 0.560 (small dataset). From visual results, it is evident that large datasets assist models in recognizing dwellings with limited ontology (as shown in the fourth example in Figure 7) or those that resemble the background (as depicted in the third example in Figure 4).

When it comes to the various performances among the models, the SAM adapter emerges as the top-performing model across all selected study sites when trained on both sizes of datasets. Its consistently high IoU values of refugee dwellings in all camps make it a promising choice for this extraction task. The FPN-MiT model also shows promising results, outperforming the remaining five semantic segmentation models in most cases. Unet-ResNet101 and Unet-MobileV3L show moderate performance. Models with ResNet34 as their backbone, both in FPN and Unet, tend to have reduced performance compared to other backbones. From the visual data, it is clear that with the same data size, SAM-Adapter performs more consistently across general built-up structures (as depicted in all examples of Figure 7), rare built-up varieties (as illustrated in the second and fourth examples of Figure 6), and structures obscured by vegetation (as in the third example of Figure 5). However, SAM-Adapter sometimes mistakenly classifies parts of the background as dwellings, as shown in the third example in Figure 6.

\begin{table}[ht]
    \centering
    \scriptsize
    \caption{Accuracy assessment results of Kutupalong and Nduta refugee camps.}
    \label{tab:table2}
    
    \begin{tabular}{cccccccccc}
        \toprule
        \textbf{Model} & \textbf{Data size} & \multicolumn{4}{c}{\textbf{Kutupalong}} & \multicolumn{4}{c}{\textbf{Nduta}}  \\
        
        \cmidrule(lr){3-6} \cmidrule(lr){7-10} 
         &  & IoU & F1 & Prec.& Rec. & IoU & F1 & Prec. & Rec. \\
         
        \midrule
        FPN-MiT & Large & \textbf{0.666} & 0.800 & 0.762 & 0.841 & \textbf{0.558} & 0.717 & 0.781 & 0.662 \\
         & Small & 0.656 & 0.792 & 0.825 & 0.763 & 0.456 & 0.627 & 0.748 & 0.539 \\
         
        \midrule
        FPN-MobileV3 & Large & 0.602 & 0.751 & 0.818 & 0.695 & 0.441 & 0.612 & 0.850 & 0.478 \\
         & Small & 0.423 & 0.594 & 0.909 & 0.441 & 0.397 & 0.568 & 0.847 & 0.427 \\
         
        \midrule
        FPN-ResNet34 & Large & 0.594 & 0.745 & 0.808 & 0.692 & 0.490 & 0.658 & 0.809 & 0.555 \\
         & Small & 0.587 & 0.740 & 0.805 & 0.684 & 0.201 & 0.286 & 0.759 & 0.176 \\
         
        \midrule
        Unet-ResNet101 & Large & 0.574 & 0.729 & 0.669 & 0.802 & 0.515 & 0.680 & 0.808 & 0.586 \\
         & Small & 0.519 & 0.684 & 0.764 & 0.618 & 0.394 & 0.570 & 0.702 & 0.480 \\
         
        \midrule
        Unet-MobileV3 & Large & 0.608 & 0.757 & 0.815 & 0.706 & 0.508 & 0.673 & 0.838 & 0.563 \\
        & Small & 0.593 & 0.745 & 0.820 & 0.682 & 0.404 & 0.575 & 0.669 & 0.505 \\
         
        \midrule
        Unet-ResNet34 & Large & 0.606 & 0.755 & 0.695 & 0.826 & 0.340 & 0.508 & 0.870 & 0.359 \\
         & Small & 0.601 & 0.750 & 0.790 & 0.715 & 0.268 & 0.423 & 0.721 & 0.300 \\
         
        \midrule
       SAM-Adapter & Large & \textbf{0.733} & 0.846 & 0.879 & 0.815 & \textbf{0.639} & 0.780 & 0.793 & 0.767 \\
         & Small & \textbf{0.710} & 0.831 & 0.810 & 0.852 & \textbf{0.618} & 0.764 & 0.793 & 0.737 \\
         
        \midrule
        SAM & noFT & 0.453 & 0.623 & 0.483 & 0.878 & 0.041 & 0.079 & 0.041 & 0.745 \\
        
        \bottomrule
    \end{tabular}
    \captionsetup{justification=centering}
    
\end{table}

\begin{table}[ht]
    \centering
    \scriptsize
    \caption{Accuracy assessment results of Dagaheley, Djibo, and Minawao refugee camps.}
    \label{tab:table3}
    
    \begin{tabular}{cccccccccccccc}
        \toprule
        \textbf{Model} & \textbf{Data size} & \multicolumn{4}{c}{\textbf{Dagaheley}} & \multicolumn{4}{c}{\textbf{Djibo}} & \multicolumn{4}{c}{\textbf{Minawao}} \\
        \cmidrule(lr){3-6} \cmidrule(lr){7-10} \cmidrule(lr){11-14}
         &  & IoU & F1 & Prec.& Rec. & IoU & F1 & Prec. & Rec. & IoU & F1 & Prec. & Rec. \\
         
        \midrule
        FPN-MiT & Large & \textbf{0.523} & 0.687 & 0.784 & 0.612 & \textbf{0.546} & 0.706 & 0.762 & 0.658 & \textbf{0.515} & 0.680 & 0.648 & 0.716 \\
         & Small & 0.297 & 0.458 & 0.429 & 0.490 & 0.284 & 0.443 & 0.365 & 0.563 & 0.194 & 0.326 & 0.621 & 0.221 \\
         
        \midrule
        FPN-MobileV3L & Large & 0.465 & 0.635 & 0.835 & 0.513 & 0.461 & 0.631 & 0.844 & 0.504 & 0.351 & 0.519 & 0.821 & 0.380 \\
         & Small & 0.251 & 0.402 & 0.564 & 0.312 & \textbf{0.304} & 0.466 & 0.610 & 0.377 & \textbf{0.195} & 0.326 & 0.668 & 0.216 \\
         
        \midrule
        FPN-ResNet34 & Large & 0.351 & 0.519 & 0.803 & 0.384 & 0.293 & 0.453 & 0.857 & 0.308 & 0.158 & 0.273 & 0.769 & 0.166 \\
         & Small & 0.138 & 0.239 & 0.721 & 0.143 & 0.107 & 0.267 & 0.170 & 0.614 & 0.139 & 0.180 & 0.114 & 0.429 \\
         
        \midrule
        Unet-ResNet101 & Large & 0.505 & 0.671 & 0.670 & 0.672 & 0.455 & 0.626 & 0.749 & 0.537 & 0.261 & 0.414 & 0.413 & 0.415 \\
         & Small & 0.140 & 0.245 & 0.146 & 0.758 & 0.118 & 0.301 & 0.215 & 0.502 & 0.121 & 0.245 & 0.153 & 0.622 \\
         
        \midrule
        Unet-MobileV3 & Large & \textbf{0.557} & 0.715 & 0.657 & 0.785 & 0.453 & 0.623 & 0.705 & 0.559 & 0.278 & 0.435 & 0.857 & 0.291 \\
         & Small & 0.159 & 0.274 & 0.265 & 0.284 & 0.128 & 0.302 & 0.424 & 0.308 & 0.129 & 0.228 & 0.424 & 0.156 \\
         
        \midrule
        Unet-ResNet34 & Large & 0.432 & 0.590 & 0.793 & 0.504 & 0.328 & 0.553 & 0.861 & 0.364 & 0.340 & 0.461 & 0.861 & 0.364 \\
         & Small & 0.129 & 0.229 & 0.141 & 0.612 & 0.158 & 0.272 & 0.144 & 0.252 & 0.144 & 0.252 & 0.164 & 0.541 \\
         
        \midrule
        SAM-Adapter & Large & \textbf{0.619} & 0.765 & 0.793 & 0.738 & \textbf{0.657} & 0.749 & 0.783 & 0.693 & \textbf{0.583} & 0.736 & 0.779 & 0.698 \\
         & Small & \textbf{0.560} & 0.703 & 0.626 & 0.842 & \textbf{0.588} & 0.741 & 0.769 & 0.737 & \textbf{0.571} & 0.727 & 0.699 & 0.757 \\
         
        \midrule
        SAM & noFT & 0.160 & 0.231 & 0.156 & 0.795 & 0.093 & 0.170 & 0.093 & 0.735 & 0.067 & 0.125 & 0.068 & 0.810 \\
         & SR-noFT & 0.219 & 0.360 & 0.231 & 0.809 & 0.104 & 0.189 & 0.107 & 0.833 & 0.067 & 0.125 & 0.068 & 0.810 \\
        \bottomrule
    \end{tabular}
    \captionsetup{justification=centering}
    
\end{table}

Regarding location-based differences, Kutupalong emerges as a preferred site for the majority of models, often recording their peak performances there. The IoU scores are highest at 0.733 and 0.710 for large and small datasets, respectively. This may be attributed to Kutupalong's higher satellite image resolution (0.1 m), capturing more intricate details of building roofs. Conversely, while Nduta's training patch data aligns more with Kutupalong than other sites, the IoU scores are 0.639 and 0.618 for the respective datasets. Notably, Nduta's building structures are simpler compared to Kutupalong. The findings indicate that higher image resolutions typically yield better segmentation results, as mentioned in ~\cite{Osco2023Segment}. Additionally, Nduta's results might be skewed by vegetation coverage, as illustrated in Figure \ref{fig:fig5}. Both the SAM-Adapter (with a small dataset) and FPN-MiT struggle to identify buildings obscured by vegetation. Among the other camps with comparable training patches, Minawao and Dagahaley pose significant segmentation challenges, generally yielding lower accuracy metrics. For instance, Minawao's IoU scores for six of the models using small datasets are below 0.2. Despite Dagahaley's superior training patch numbers and resolution compared to Djibo, the SAM-Adapter performs better in Djibo. Other models display variable results, yet most excel in the Dagahaley. When considering factors like building structure and background, the data from Dagahaley seem more favorable than that from Djibo, as evidenced by the performance of other models. This observation could potentially be attributed to the inherent tendencies of the SAM model, which requires more research in the future. 

\begin{figure}[ht]
    \centering
    \includegraphics[width=0.9\linewidth]{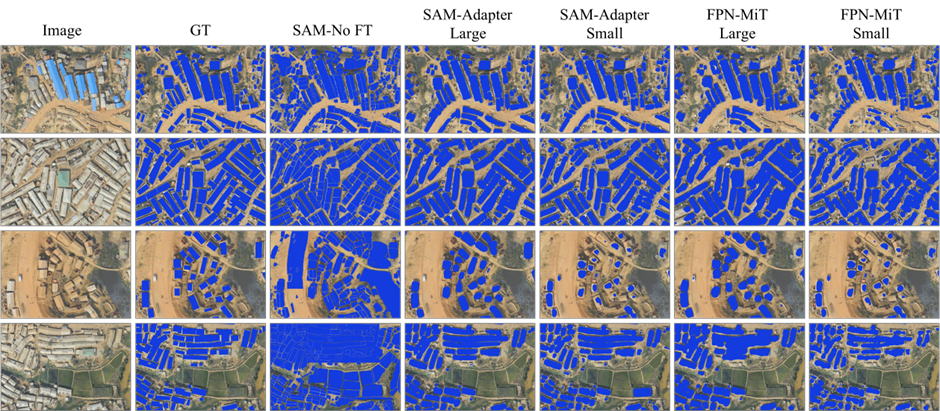}
    \caption{Qualitative results of predicted masks from SAM-Adapter, SAM without fine-tuning, and FPN-MiT in Kutupalong refugee camp.}
    \label{fig:fig4}
\end{figure}
\vspace{-1cm} 

\begin{figure}[h]
    \centering
    \includegraphics[width=0.9\linewidth]{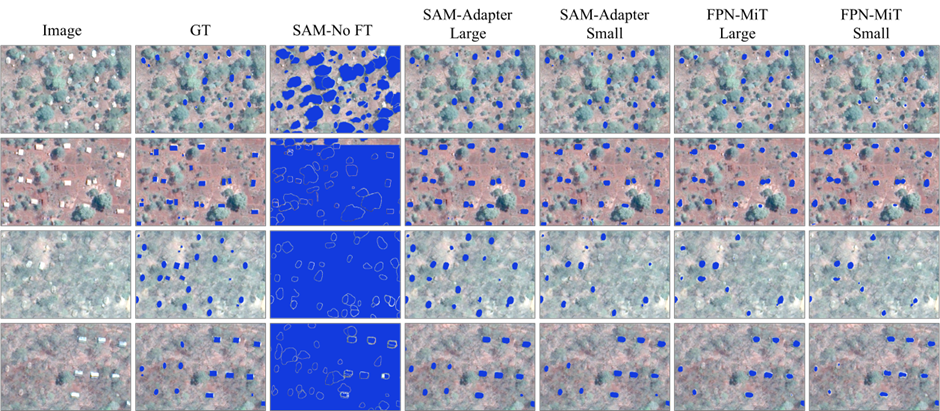}
    \caption{Qualitative results of predicted masks from SAM-Adapter, SAM without fine-tuning, and FPN-MiT in Nduta refugee camp.}
    \label{fig:fig5}
\end{figure}
\vspace{-1cm} 

\begin{figure}[h]
    \centering
    \includegraphics[width=0.9\linewidth]{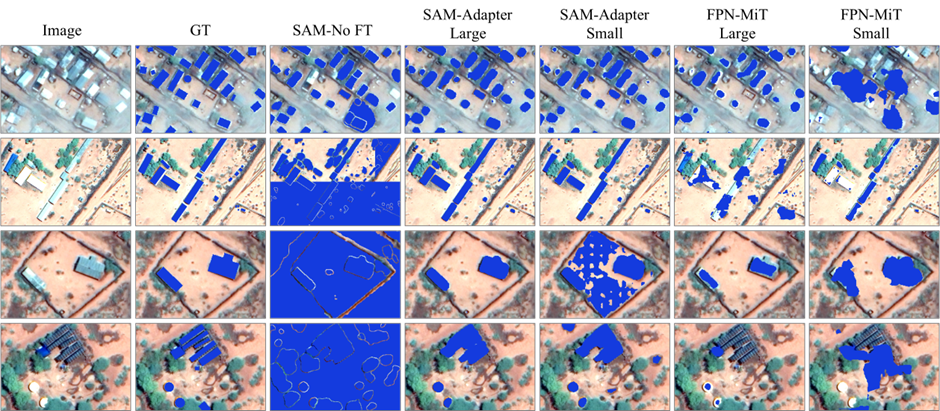}
    \caption{Qualitative results of predicted masks from SAM-Adapter, SAM without fine-tuning, and FPN-MiT in Dagahaley refugee camp.}
    \label{fig:fig6}
\end{figure}
\vspace{-1cm} 
\clearpage 

\begin{figure}[ht]
    \centering
    \includegraphics[width=0.9\linewidth]{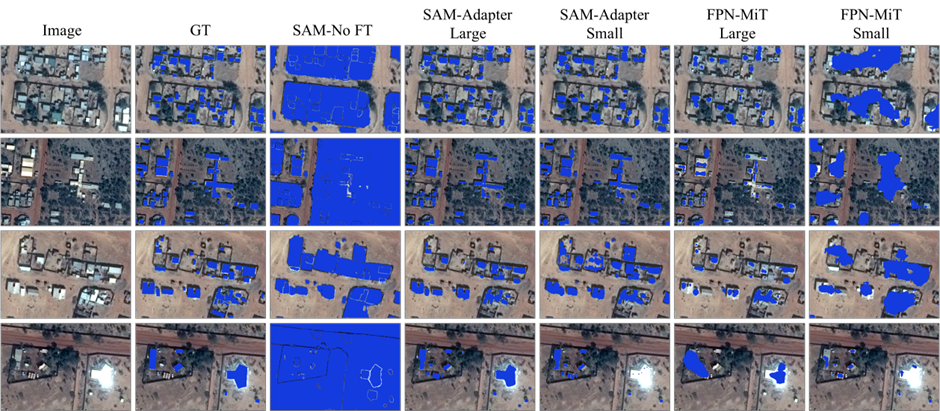}
    \caption{Qualitative results of predicted masks from SAM-Adapter, SAM without fine-tuning, and FPN-MiT in Djibo refugee camp.}
    \label{fig:fig7}
\end{figure}
\vspace{-1cm} 

\begin{figure}[h]
    \centering
    \includegraphics[width=0.9\linewidth]{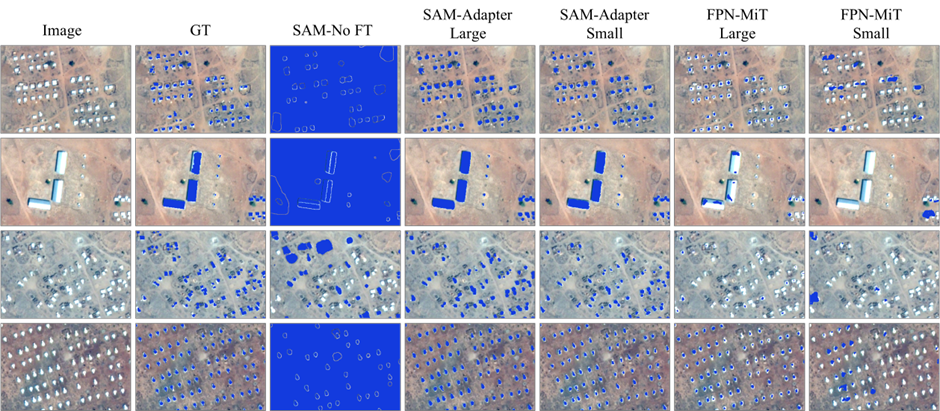}
    \caption{Qualitative results of predicted masks from SAM-Adapter, SAM without fine-tuning, and FPN-MiT in Minawao refugee camp.}
    \label{fig:fig8}
\end{figure}


\subsection{Analyzing the influence of upscaling}
Table \ref{tab:table4} presents the accuracy assessment results for the Dagahaley, Djibo, and Minawao refugee camps.  The top IoU results for each model, site, and data size are marked in bold.

\begin{table}[H]
    \centering
    \scriptsize
    \caption{Accuracy assessment results of SAM-Adapter and FPN-MiT, trained on upscaled Train-Small datasets.}
    \label{tab:table4}
    
    \begin{tabular}{cccccccccccccc}
        \toprule
        \textbf{Model} & \textbf{Dataset} & \multicolumn{4}{c}{\textbf{Dagaheley}} & \multicolumn{4}{c}{\textbf{Djibo}} & \multicolumn{4}{c}{\textbf{Minawao}} \\
        \cmidrule(lr){3-6} \cmidrule(lr){7-10} \cmidrule(lr){11-14}
         &  & IoU & F1 & Prec.& Rec. & IoU & F1 & Prec. & Rec. & IoU & F1 & Prec. & Rec. \\
         
        \midrule
        FPN-MiT & Aug. & 0.297 & 0.458 & 0.429 & 0.490 & 0.284 & 0.443 & 0.365 & 0.563 & 0.194 & 0.326 & 0.621 & 0.221 \\
         & Aug.+Nearest & 0.553 & 0.712 & 0.683 & 0.743 & 0.547 & 0.707 & 0.755 & 0.665 & 0.347 & 0.515 & 0.817 & 0.376 \\
         & Aug.+Bilinear & \textbf{0.561} & 0.718 & 0.709 & 0.728 & \textbf{0.561} & 0.719 & 0.754 & 0.686 & 0.366 & 0.536 & 0.804 & 0.402 \\
         & Aug.+EDSR & 0.556 & 0.714 & 0.692 & 0.739 & 0.559 & 0.717 & 0.744 & 0.693 & 0.404 & 0.576 & 0.782 & 0.456 \\
         
        \midrule
        SAM-Adapter & Aug. & 0.618 & 0.764 & 0.859 & 0.687 & 0.587 & 0.740 & 0.822 & 0.672 & 0.581 & 0.735 & 0.791 & 0.686 \\
         & Nearest & 0.619 & 0.769 & 0.750 & 0.790 & 0.631 & 0.774 & 0.747 & 0.802 & 0.457 & 0.627 & 0.490 & 0.872 \\
         & Bilinear & 0.643 & 0.783 & 0.786 & 0.780 & 0.651 & 0.789 & 0.789 & 0.788 & 0.565 & 0.722 & 0.639 & 0.830 \\
         & EDSR & \textbf{0.660} & 0.795 & 0.803 & 0.788 & \textbf{0.653} & 0.790 & 0.769 & 0.813 & \textbf{0.613} & 0.760 & 0.754 & 0.767 \\
        \hline
        \bottomrule
        
    \end{tabular} 
    
\end{table}
\vspace{-1cm}

\begin{figure}[h]
    \centering
    \includegraphics[width=0.9\linewidth]{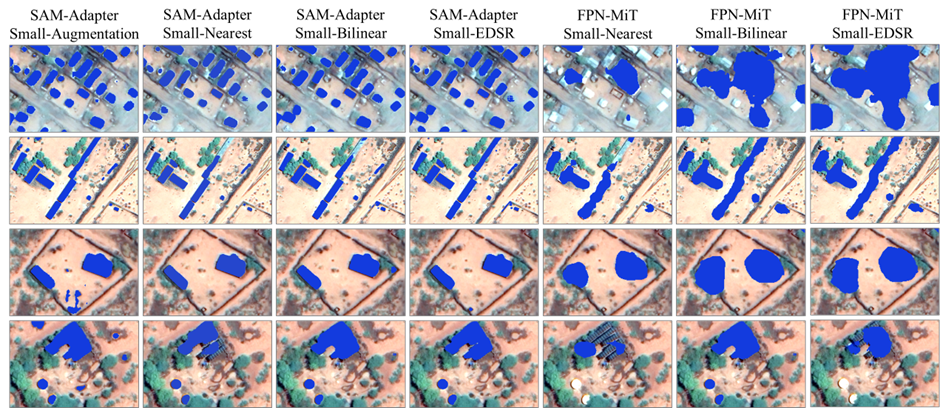}
    \caption{Qualitative results of predicted masks from SAM-Adapter and FPN-MiT when using Train-Small datasets upscaled by three approaches in the Dagahaley refugee camp.}
    \label{fig:fig9}
\end{figure}
\vspace{-1cm}

\begin{figure}[h]
    \centering
    \includegraphics[width=0.9\linewidth]{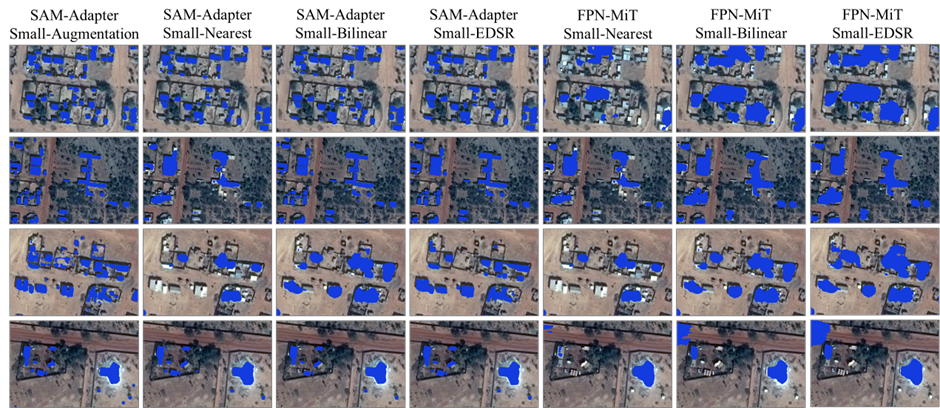}
    \caption{Qualitative results of predicted masks from SAM-Adapter and FPN-MiT when using Train-Small datasets upscaled by three approaches in the Djibo refugee camp.}
    \label{fig:fig10}
\end{figure}
\vspace{-1cm}

\begin{figure}[h]
    \centering
    \includegraphics[width=0.9\linewidth]{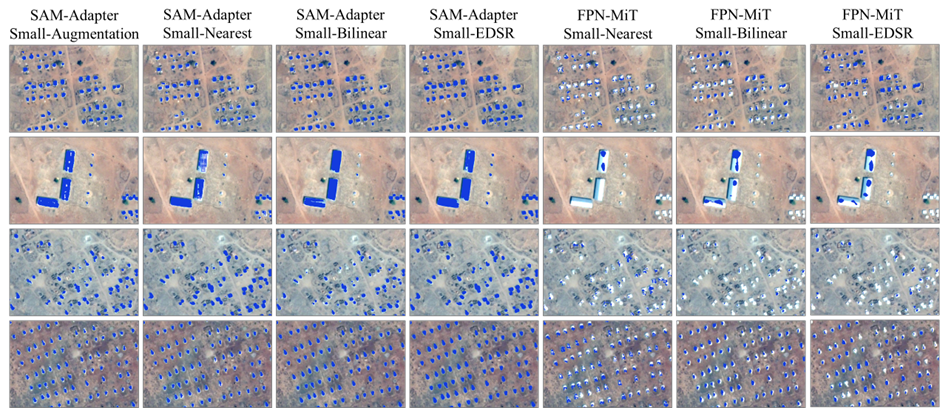}
    \caption{Qualitative results of predicted masks from SAM-Adapter and FPN-MiT when using Train-Small datasets upscaled by three approaches in the Minawao refugee camp.}
    \label{fig:fig11}
\end{figure}
\clearpage

\begin{figure}[h]
    \centering
    \includegraphics[width=0.6\linewidth]{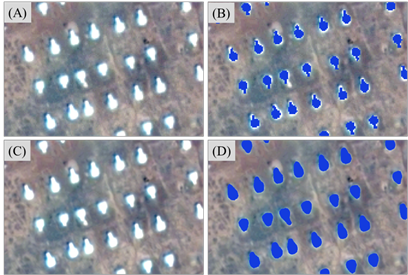}
    \caption{An example in the Minawao refugee camp showcases the influence of upscaling by SR models on the performance of SAM-Adapter. (A) Original image; (B) Ground truth; (C) Upscaled image; (D) Predicted masks from SAM-Adapter, which are smoother than ground truth labels.}
    \label{fig:fig12}
\end{figure}

Overall, the EDSR model outperforms other upscaling methods in terms of IoU and F1 across all three sites, making it the most effective augmentation for the SAM-Adapter model in these scenarios. Besides, it can be observed that SAM-Adapter tends to have higher IoU values across most upscaling methods and sites when compared to FPN-MiT. For instance, in Dagahaley, upscaled by EDSR, the SAM-Adapter achieves an IoU of 0.660, which is noticeably higher than the 0.556 IoU achieved by FPN-MiT using Aug.+EDSR. For FPN-MiT, upscaling significantly improves the performance, as seen in the Dagahaley site where IoU jumps from 0.297 with standard augmentation to 0.556 when combined with EDSR upscaling. For SAM-Adapter, the application of various upscaling methods and data augmentation yields mixed outcomes. In Djibo, for instance, the model attains an IoU of 0.588 without augmentation, but this slightly decreases to 0.587 when augmentation is introduced. However, in Dagahaley, the IoU value rises from 0.560 to 0.618 with the application of augmentation. For Minawao with the Nearest interpolation method, despite a lower IoU and F1 score, the model achieves an impressively high Recall of 0.872, suggesting that while it identifies most of the buildings, it also makes more false positive errors (lower Precision of 0.49).

Figure \ref{fig:fig9} through 11 show visual representations of predicted masks under different settings. From the observations, it is evident that the visual outcomes have improved for most cases compared to models trained on small datasets without upscaling. SAM-Adapter's predicted masks closely align with the ground truth. However, the predictions from FPN-MiT remain unsatisfactory, showing significant over-segmentation (as seen in Figure \ref{fig:fig9}) and under-segmentation (as seen in Figure \ref{fig:fig11}).

Figure \ref{fig:fig12} showcases a sample of the predicted masks produced by the SAM-Adapter when trained with smaller datasets and upscaled using the SR model. Compared to the ground truth, these masks exhibit smoother contours, potentially making them more suitable for shape analysis once transformed into polygons.

\subsection{Peak results from the first epoch}
Table \ref{tab:table5} presents the results of the accuracy assessment in each training epoch of SAM-Adapter for the Dagahaley, Djibo, and Minawao refugee camps in different settings. The top IoU results for each model, study site, data size, and upscaling approach are marked in bold.
From these results, it can be found that for all of the Dagahaley, Djibo, and Minawao datasets, the upscaling techniques achieve their peak performance in the first epoch. For Dagahaley (Small), EDSR consistently emerges as the top-performing method across all epochs, highlighting its efficacy. In contrast, the Nearest interpolation method shows significant fluctuations, with a notable dip in the third epoch. Turning to Djibo (Small), the Nearest interpolation method takes an initial lead in the first epoch but fails to sustain this advantage. Lastly, in the Minawao (Small) dataset, the Nearest method exhibits major performance fluctuations, especially the drastic dip in the second epoch. Yet again, EDSR maintains the best performance. In an overarching observation, the superior performance of EDSR stands out, while the variability of the Nearest method remains a cause for concern. 

\begin{figure}[h]
    \centering
    \includegraphics[width=1\linewidth]{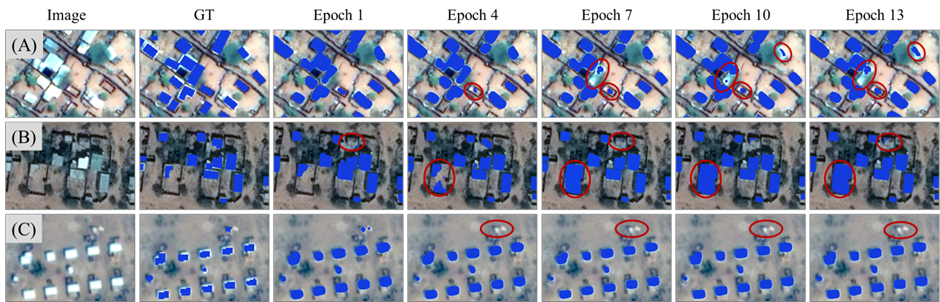}
    \caption{Qualitative results of predicted masks from SAM-Adapter, trained with small datasets upscaled by EDSR model across various training epochs, in (A) Dagahaley, (B) Djibo, and (C) Minawao refugee camps. Some errors in predicted masks are indicated by red circles.}
    \label{fig:fig13}
\end{figure}

Figure \ref{fig:fig13} displays the visual results of the SAM-Adapter predicted masks, trained with the Train-Small dataset upscaled by the EDSR model across various training epochs. Some errors in predicted masks are indicated by red circles. It can be observed that the model trained with small datasets upscaled by the EDSR can gradually produce more false positive pixels (as shown in Figure \ref{fig:fig13} (B)) and lose true positive pixels (as shown in Figure \ref{fig:fig13} (A)).

\begin{table}[ht!]
    \centering
    \scriptsize
    \caption{IoU values of predicted masks from SAM-Adapter at different training epochs when using different datasets for the Dagaheley, Djibo, and Minawao refugee camps.}
    \setlength{\tabcolsep}{4pt} 
    \label{tab:table5}
    
    \begin{tabular}{|c|c|ccccc|c|ccccc|c|ccccc|}
        \hline
        EP. & \multicolumn{6}{c|}{Dagaheley} & \multicolumn{6}{c|}{Djibo} & \multicolumn{6}{c|}{Minawao} \\ 
        \hline
        & \multicolumn{1}{c|}{Large} & \multicolumn{5}{c|}{Small} & \multicolumn{1}{c|}{Large} & \multicolumn{5}{c|}{Small} & \multicolumn{1}{c|}{Large} & \multicolumn{5}{c|}{Small} \\ 
        & & Original & Aug. & Nearest & Bilinear & EDSR & & Original & Aug. & Nearest & Bilinear & EDSR & & Original & Aug. & Nearest & Bilinear & EDSR \\
        \hline
        1 & 0.623 & 0.566 & 0.600 & \textbf{0.619} & 0.643 & \textbf{0.660} & 0.543 & 0.507 & 0.516 & \textbf{0.631} & 0.651 & \textbf{0.653} & 0.560 & 0.519 & 0.563 & \textbf{0.457} & 0.565 & \textbf{0.613} \\
        2 & 0.637 & 0.608 & 0.602 & 0.583 & 0.635 & 0.643 & 0.646 & 0.567 & 0.579 & 0.621 & 0.648 & 0.645 & 0.558 & 0.531 & 0.572 & 0.082 & 0.526 & 0.605 \\
        3 & 0.631 & \textbf{0.618} & 0.599 & 0.474 & 0.641 & 0.655 & 0.657 & 0.569 & 0.568 & 0.618 & 0.612 & \textbf{0.651} & 0.581 & 0.544 & 0.576 & 0.309 & 0.427 & 0.607 \\
        4 & 0.619 & 0.612 & \textbf{0.618} & 0.578 & 0.614 & 0.649 & 0.662 & \textbf{0.588} & 0.562 & 0.575 & 0.639 & \textbf{0.639} & 0.582 & 0.543 & 0.580 & 0.210 & 0.510 & 0.600 \\
        5 & 0.637 & 0.603 & 0.608 & 0.588 & 0.618 & 0.641 & \textbf{0.664} & 0.574 & \textbf{0.587} & 0.584 & 0.638 & \textbf{0.653} & 0.582 & 0.566 & 0.571 & 0.248 & 0.537 & 0.604 \\
        6 & 0.618 & 0.601 & 0.597 & 0.583 & 0.609 & 0.643 & 0.661 & 0.559 & 0.581 & 0.580 & 0.619 & \textbf{0.644} & 0.581 & 0.561 & 0.520 & 0.107 & 0.472 & 0.583 \\
        7 & 0.630 & 0.594 & 0.586 & 0.567 & 0.603 & 0.631 & 0.644 & 0.525 & 0.580 & 0.592 & 0.633 & \textbf{0.638} & 0.579 & 0.564 & 0.559 & 0.232 & 0.499 & 0.577 \\
        8 & 0.633 & 0.602 & 0.586 & 0.535 & 0.520 & 0.627 & 0.662 & 0.539 & 0.581 & 0.569 & 0.623 & \textbf{0.631} & 0.582 & 0.555 & 0.568 & 0.116 & 0.514 & 0.595 \\
        9 & 0.638 & 0.601 & 0.596 & 0.529 & 0.511 & 0.622 & 0.654 & 0.502 & 0.572 & 0.578 & 0.618 & \textbf{0.641} & 0.582 & 0.559 & 0.568 & 0.235 & 0.458 & 0.586 \\
        10 & 0.633 & 0.604 & 0.594 & 0.496 & 0.508 & 0.627 & 0.647 & 0.517 & 0.586 & 0.563 & 0.615 & \textbf{0.635} & 0.581 & 0.558 & 0.575 & 0.318 & 0.551 & 0.590 \\
        11 & 0.637 & 0.598 & 0.595 & 0.496 & 0.467 & 0.619 & 0.638 & 0.512 & 0.566 & 0.570 & 0.608 & \textbf{0.636} & 0.581 & 0.569 & 0.574 & 0.246 & 0.552 & 0.590 \\
        12 & 0.639 & 0.595 & 0.601 & 0.535 & 0.518 & 0.619 & 0.637 & 0.536 & 0.576 & 0.571 & 0.603 & \textbf{0.634} & 0.582 & 0.569 & 0.573 & 0.192 & 0.554 & 0.595 \\
        13 & 0.637 & 0.599 & 0.596 & 0.515 & 0.487 & 0.608 & 0.650 & 0.509 & 0.579 & 0.565 & 0.597 & \textbf{0.625} & 0.580 & \textbf{0.571} & \textbf{0.581} & 0.198 & 0.560 & 0.593 \\
        \hline
    \end{tabular}
\end{table}

\section{Conclusions and Outlooks}
\label{sec:others}
In conclusion, this study has demonstrated the key role of SAM-Adapter in addressing the complex refugee dwelling extraction task from high-resolution satellite imagery. Based on the results of this research, we can draw the following conclusions.

Firstly, SAM-Adapter is a highly effective tool for semantic segmentation tasks, especially when faced with limited pixel-wise labeled data. Its adaptability and remarkable performance across diverse refugee camps highlight its potential in building extraction within refugee camps for humanitarian operations.

Secondly, the significance of upscaling methods on model performance is profoundly essential. The application of SR model techniques such as EDSR for image data significantly enhances the quality of the predictions, making them more reliable for practical applications.

Third, our study reaffirms that higher image resolutions contribute to better segmentation outcomes (e.g., Kutupalong), while variations in building structures and background complexity pose unique challenges that must be considered in the application of SAM-Adapter or other SAM-based models.

Furthermore, our research hints at intriguing possibilities, such as the model's rapid convergence in the first training epoch when using upscaled image data, opening avenues for future exploration.

Overall, there's significant room for enhancement of utlizing SAM or other foundations models. Firstly, enhancing the fine-tuning speed of SAM is essential. There have been recent advancements in SAM models, with the introduction of versions like Faster SAM ~\cite{Zhang2023Faster}. Integrating these more efficient models and fine-tuning them with specific data could be valuable.

Secondly, while the SAM-Adapter's architecture is uncomplicated and easy to implement, there's some room for modifications. One straightforward change could be decreasing the number of adapters.
Thirdly, there's a potential in devising more efficient sampling strategies. While our study employed a rudimentary sampling approach, it underscored SAM's potential for this extraction task. Optimized sampling methods might further boost SAM's performance.

Last but not least, creating SR models tailored for satellite imagery that can surpass a spatial resolution of 0.1 m could be significant for building extraction tasks. Drawing from the findings of this study and other studies like ~\cite{Osco2023Segment}, the critical role of resolution becomes evident.



\begin{thebibliography}{9}

\bibitem{UNHCR2020Sustainable}
UNHCR,
"The Sustainable Development Goals and the Global Compact on Refugees",
\textit{}, vol. , no. , pp. , 2020.

\bibitem{Çelik2012Humanitarian}
Çelik, Melih, Ergun, Özlem, Johnson, Ben, Keskinocak, Pınar, Lorca, Álvaro, Pekgün, Pelin, Swann, Julie,
"Humanitarian logistics",
\textit{}, vol. , no. , pp. 18--49, 2012.

\bibitem{Checchi2013Validity}
Checchi, Francesco, Stewart, Barclay T., Palmer, Jennifer J., Grundy, Chris,
"Validity and feasibility of a satellite imagery-based method for rapid estimation of displaced populations",
\textit{International Journal of Health Geographics}, vol. 12, no. , pp. , 2013.

\bibitem{Spröhnle2014Earth}
Spröhnle, Kristin, Tiede, Dirk, Schoepfer, Elisabeth, Füreder, Petra, Svanberg, Anna, Rost, Torbjörn,
"Earth observation-based dwelling detection approaches in a highly complex refugee camp environment - A comparative study",
\textit{Remote Sensing}, vol. 6, no. 10, pp. 9277--9297, 2014.

\bibitem{Füreder2014Object-based}
Füreder, Petra, Tiede, Dirk, Lüthje, Fritjof, Lang, Stefan,
"Object-based dwelling extraction in refugee/IDP camps–challenges in an operational mode",
\textit{South-Eastern European Journal of Earth Observation and Geomatics}, vol. 3, no. 2S, pp. 539--544, 2014.

\bibitem{Lang2020Earth}
Lang, Stefan, Füreder, Petra, Riedler, Barbara, Wendt, Lorenz, Braun, Andreas, Tiede, Dirk, Schoepfer, Elisabeth, Zeil, Peter, Spröhnle, Kristin, Kulessa, Kerstin, Rogenhofer, Edith, Bäuerl, Magdalena, Öze, Alexander, Schwendemann, Gina, Hochschild, Volker,
"Earth observation tools and services to increase the effectiveness of humanitarian assistance",
\textit{European Journal of Remote Sensing}, vol. 53, no. sup2, pp. 67--85, 2020.

\bibitem{krafft2016template}
Krafft, Pascal, Tiede, D, F{\"u}reder, P,
"Template matching to support earth observation based refugee camp analysis in obia workflows-creation and evaluation of a dwelling template library for improving dwelling extraction within an object-based framework",
\textit{}, vol. , no. , pp. , 2016.

\bibitem{Tiede2017Stratified}
Tiede, Dirk, Krafft, Pascal, Füreder, Petra, Lang, Stefan,
"Stratified template matching to support refugee camp analysis in OBIA workflows",
\textit{Remote Sensing}, vol. 9, no. 4, pp. , 2017.

\bibitem{Tiede2013Automated}
Tiede, Dirk, Füreder, Petra, Lang, Stefan, Hölbling, Daniel, Zeil, Peter,
"Automated analysis of satellite imagery to provide information products for humanitarian relief operations in refugee camps -from scientific development towards operational services",
\textit{Photogrammetrie, Fernerkundung, Geoinformation}, vol. 2013, no. 3, pp. 185--195, 2013.

\bibitem{Tiede2010Transferability}
Tiede, D., Lang, S, Hölbling, D., Füreder, P.,
"Transferability of obia rulesets for idp camp analysis in darfur",
\textit{Geobia}, vol. 2006, no. , pp. , 2010.

\bibitem{Ghorbanzadeh2018Dwelling}
Ghorbanzadeh, Omid, Tiede, Dirk, Dabiri, Zahra, Sudmanns, Martin, Lang, Stefan,
"Dwelling extraction in refugee camps using CNN - First experiences and lessons learnt",
\textit{International Archives of the Photogrammetry, Remote Sensing and Spatial Information Sciences - ISPRS Archives}, vol. 42, no. 1, pp. 161--166, 2018.

\bibitem{Quinn2018Humanitarian}
Quinn, John A., Nyhan, Marguerite M., Navarro, Celia, Coluccia, Davide, Bromley, Lars, Luengo-Oroz, Miguel,
"Humanitarian applications of machine learning with remote-sensing data: Review and case study in refugee settlement mapping",
\textit{Philosophical Transactions of the Royal Society A: Mathematical, Physical and Engineering Sciences}, vol. 376, no. 2128, pp. , 2018.

\bibitem{Tiede2021Mask}
Tiede, Dirk, Schwendemann, Gina, Alobaidi, Ahmad, Wendt, Lorenz, Lang, Stefan,
"Mask R-CNN-based building extraction from VHR satellite data in operational humanitarian action: An example related to Covid-19 response in Khartoum, Sudan",
\textit{Transactions in GIS}, vol. 25, no. 3, pp. 1213--1227, 2021.

\bibitem{Gella2022Mapping}
Gella, Getachew Workineh, Wendt, Lorenz, Lang, Stefan, Tiede, Dirk, Hofer, Barbara, Gao, Yunya, Braun, Andreas,
"Mapping of Dwellings in IDP/Refugee Settlements from Very High-Resolution Satellite Imagery Using a Mask Region-Based Convolutional Neural Network",
\textit{Remote Sensing}, vol. 14, no. 3, pp. , 2022.

\bibitem{Gao2022Comparing}
Gao, Yunya, Lang, Stefan, Tiede, Dirk, Gella, Getachew Workineh, Wendt, Lorenz,
"Comparing the robustness of U-Net, LinkNet, and FPN towards label noise for refugee dwelling extraction from satellite imagery",
\textit{2022 IEEE Global Humanitarian Technology Conference, GHTC 2022}, vol. , no. , pp. 88--94, 2022.

\bibitem{Gella2023Spatially}
Gella, Getachew Workineh, Tiede, Dirk, Lang, Stefan, Wendit, Lorenz, Gao, Yunya,
"Spatially transferable dwelling extraction from Multi-Sensor imagery in IDP/Refugee Settlements: A meta-Learning approach",
\textit{International Journal of Applied Earth Observation and Geoinformation}, vol. 117, no. , pp. 103210, 2023.

\bibitem{Ji2023Segment}
Ji, Wei, Li, Jingjing, Bi, Qi, Liu, Tingwei, Li, Wenbo, Cheng, Li,
"Segment anything is not always perfect: An investigation of sam on different real-world applications",
\textit{}, vol. , no. , pp. , 2024.

\bibitem{Kirillov2023Segment}
Kirillov, Alexander, Mintun, Eric, Ravi, Nikhila, Mao, Hanzi, Rolland, Chloe, Gustafson, Laura, Xiao, Tete, Whitehead, Spencer, Berg, Alexander C, Lo, Wan-Yen, others,
"Segment anything",
\textit{Proceedings of the IEEE/CVF International Conference on Computer Vision}, vol. , no. , pp. 4015--4026, 2023.

\bibitem{Zhang2023Survey}
Zhang, Chaoning, Puspitasari, Fachrina Dewi, Zheng, Sheng, Li, Chenghao, Qiao, Yu, Kang, Taegoo, Shan, Xinru, Zhang, Chenshuang, Qin, Caiyan, Rameau, Francois, others,
"A survey on segment anything model (sam): Vision foundation model meets prompt engineering",
\textit{arXiv preprint arXiv:2306.06211}, vol. , no. , pp. , 2023.

\bibitem{Ren2023Segment}
Ren, Simiao, Luzi, Francesco, Lahrichi, Saad, Kassaw, Kaleb, Collins, Leslie M, Bradbury, Kyle, Malof, Jordan M,
"Segment anything, from space?",
\textit{Proceedings of the IEEE/CVF Winter Conference on Applications of Computer Vision}, vol. , no. , pp. 8355--8365, 2024.

\bibitem{Chen2022Vision}
Chen, Zhe, Duan, Yuchen, Wang, Wenhai, He, Junjun, Lu, Tong, Dai, Jifeng, Qiao, Yu,
"Vision transformer adapter for dense predictions",
\textit{arXiv preprint arXiv:2205.08534}, vol. , no. , pp. , 2022.

\bibitem{Touzani2021Open}
Touzani, Samir, Granderson, Jessica,
"Open data and deep semantic segmentation for automated extraction of building footprints",
\textit{Remote Sensing}, vol. 13, no. 2578, pp. , 2021.

\bibitem{Hoffmann2021Zooming}
Hoffmann, Eike Jens, Ali, Mohsin, Zhu, Xiao Xiang,
"Zooming Into Uncertainties: Towards Fusing Multi Zoom Level Imagery for Urban Land Use Segmentation",
\textit{International Geoscience and Remote Sensing Symposium (IGARSS)}, vol. , no. , pp. 2090--2093, 2021.

\bibitem{Osco2023Segment}
Osco, Lucas Prado, Wu, Qiusheng, de Lemos, Eduardo Lopes, Gon{\c{c}}alves, Wesley Nunes, Ramos, Ana Paula Marques, Li, Jonathan, Junior, Jos{\'e} Marcato,
"The segment anything model (sam) for remote sensing applications: From zero to one shot",
\textit{International Journal of Applied Earth Observation and Geoinformation}, vol. 124, no. , pp. 103540, 2023.

\bibitem{Liu2023Grounding}
Liu, Shilong, Zeng, Zhaoyang, Ren, Tianhe, Li, Feng, Zhang, Hao, Yang, Jie, Li, Chunyuan, Yang, Jianwei, Su, Hang, Zhu, Jun,
"Grounding dino: Marrying dino with grounded pre-training for open-set object detection",
\textit{arXiv preprint arXiv:2303.05499}, vol. , no. , pp. , 2023.

\bibitem{Mo2022Review}
Mo, Yujian, Wu, Yan, Yang, Xinneng, Liu, Feilin, Liao, Yujun,
"Review the state-of-the-art technologies of semantic segmentation based on deep learning",
\textit{Neurocomputing}, vol. 493, no. , pp. 626--646, 2022.

\bibitem{Ulku2022survey}
Ulku, Irem, Akagündüz, Erdem,
"A survey on deep learning-based architectures for semantic segmentation on 2d images",
\textit{Applied Artificial Intelligence}, vol. 36, no. 1, pp. 2032924, 2022.

\bibitem{Yuan2021review}
Yuan, Xiaohui, Shi, Jianfang, Gu, Lichuan,
"A review of deep learning methods for semantic segmentation of remote sensing imagery",
\textit{Expert Systems with Applications}, vol. 169, no. November 2020, pp. 114417, 2021.

\bibitem{Huang2023On}
Huang, Yihao, Cao, Yue, Li, Tianlin, Juefei-Xu, Felix, Lin, Di, Tsang, Ivor W, Liu, Yang, Guo, Qing,
"On the robustness of segment anything",
\textit{arXiv preprint arXiv:2305.16220}, vol. , no. , pp. , 2023.

\bibitem{Zhang2023Customized}
Zhang, Kaidong, Liu, Dong,
"Customized segment anything model for medical image segmentation",
\textit{arXiv preprint arXiv:2304.13785}, vol. , no. , pp. , 2023.

\bibitem{Wu2023Medical}
Wu, Junde, Fu, Rao, Fang, Huihui, Liu, Yuanpei, Wang, Zhaowei, Xu, Yanwu, Jin, Yueming, Arbel, Tal,
"Medical sam adapter: Adapting segment anything model for medical image segmentation",
\textit{arXiv preprint arXiv:2304.12620}, vol. , no. , pp. , 2023.

\bibitem{Qiu2023Learnable}
Qiu, Zhongxi, Hu, Yan, Li, Heng, Liu, Jiang,
"Learnable ophthalmology sam",
\textit{arXiv preprint arXiv:2304.13425}, vol. , no. , pp. , 2023.

\bibitem{Gao2023DeSAM:}
Gao, Yifan, Xia, Wei, Hu, Dingdu, Gao, Xin,
"DeSAM: Decoupling Segment Anything Model for Generalizable Medical Image Segmentation",
\textit{arXiv preprint arXiv:2306.00499}, vol. , no. , pp. , 2023.

\bibitem{zhang2023text2seg}
Zhang, Jielu, Zhou, Zhongliang, Mai, Gengchen, Mu, Lan, Hu, Mengxuan, Li, Sheng,
"Text2seg: Remote sensing image semantic segmentation via text-guided visual foundation models",
\textit{arXiv preprint arXiv:2304.10597}, vol. , no. , pp. , 2023.

\bibitem{Radford2021Learning}
Radford, Alec, Kim, Jong Wook, Hallacy, Chris, Ramesh, Aditya, Goh, Gabriel, Agarwal, Sandhini, Sastry, Girish, Askell, Amanda, Mishkin, Pamela, Clark, Jack, others,
"Learning transferable visual models from natural language supervision",
\textit{}, vol. , no. , pp. 8748--8763, 2021.

\bibitem{chen2024rsprompter}
Chen, Keyan, Liu, Chenyang, Chen, Hao, Zhang, Haotian, Li, Wenyuan, Zou, Zhengxia, Shi, Zhenwei,
"RSPrompter: Learning to prompt for remote sensing instance segmentation based on visual foundation model",
\textit{IEEE Transactions on Geoscience and Remote Sensing}, vol. , no. , pp. , 2024.

\bibitem{Zhang2023Personalize}
Zhang, Renrui, Jiang, Zhengkai, Guo, Ziyu, Yan, Shilin, Pan, Junting, Dong, Hao, Gao, Peng, Li, Hongsheng,
"Personalize segment anything model with one shot",
\textit{arXiv preprint arXiv:2305.03048}, vol. , no. , pp. , 2023.

\bibitem{chen2023sam}
Chen, Tianrun, Zhu, Lanyun, Ding, Chaotao, Cao, Runlong, Wang, Yan, Li, Zejian, Sun, Lingyun, Mao, Papa, Zang, Ying,
"SAM Fails to Segment Anything?--SAM-Adapter: Adapting SAM in Underperformed Scenes: Camouflage, Shadow, Medical Image Segmentation, and More",
\textit{arXiv preprint arXiv:2304.09148}, vol. , no. , pp. , 2023.

\bibitem{UNHCR2023Inside}
UNHCR,
"Inside the world's five largest refugee camps",
\textit{United Nations High Commissioner for Refugees}, vol. , no. , pp. , 2023.

\bibitem{UNHCR2018North-West}
UNHCR,
"North-West Tanzania Nduta Refugee Camp Profile",
\textit{}, vol. , no. , pp. , 2018.

\bibitem{UNHCR2015Dagahaley}
UNHCR,
"Dagahaley Camp Profile, Dadaab Refugee Camps, Kenya",
\textit{}, vol. , no. , pp. , 2015.

\bibitem{UNHCR2020Burkina}
UNHCR,
"Burkina Faso: UNHCR condemns violence against Malian refugees",
\textit{The United Nations High Commissioner for Refugees (UNHCR)}, vol. , no. , pp. , 2020.

\bibitem{UNHCR2022Annual}
UNHCR,
"Annual Results Report 2022 Cameroon Multi Country",
\textit{The United Nations High Commissioner for Refugees (UNHCR)}, vol. , no. , pp. , 2022.

\bibitem{Milosavljević2020Automated}
Milosavljević, Aleksandar,
"Automated processing of remote sensing imagery using deep semantic segmentation: A building footprint extraction case",
\textit{ISPRS International Journal of Geo-Information}, vol. 9, no. 8, pp. 486, 2020.

\bibitem{Chen2023Large-scale}
Chen, Shenglong, Ogawa, Yoshiki, Zhao, Chenbo, Sekimoto, Yoshihide,
"Large-scale individual building extraction from open-source satellite imagery via super-resolution-based instance segmentation approach",
\textit{ISPRS Journal of Photogrammetry and Remote Sensing}, vol. 195, no. , pp. 129--152, 2023.

\bibitem{Lim2017Enhanced}
Lim, Bee, Son, Sanghyun, Kim, Heewon, Nah, Seungjun, Mu Lee, Kyoung,
"Enhanced deep residual networks for single image super-resolution",
\textit{}, vol. , no. , pp. 136--144, 2017.

\bibitem{Agarwal2021Super}
Agarwal, Vardan, Patnaik, Lipi,
"Super Resolution in OpenCV",
\textit{LearnOpenCV}, vol. , no. , pp. , 2021.

\bibitem{Zhang2018Residual}
Zhang, Yulun, Tian, Yapeng, Kong, Yu, Zhong, Bineng, Fu, Yun,
"Residual dense network for image super-resolution",
\textit{}, vol. , no. , pp. 2472--2481, 2018.

\bibitem{Ahn2018Fast}
Ahn, Namhyuk, Kang, Byungkon, Sohn, Kyung-Ah,
"Fast, accurate, and lightweight super-resolution with cascading residual network",
\textit{}, vol. , no. , pp. 252--268, 2018.

\bibitem{cresson2022sr4rs}
Cresson, R{\'e}mi,
"SR4RS: A Tool for Super Resolution of Remote Sensing Images",
\textit{Journal of Open Research Software}, vol. 10, no. 1, pp. , 2022.

\bibitem{Liu2023Explicit}
Liu, Weihuang, Shen, Xi, Pun, Chi-Man, Cun, Xiaodong,
"Explicit visual prompting for low-level structure segmentations",
\textit{}, vol. , no. , pp. 19434--19445, 2023.

\bibitem{Ramchoun2016Multilayer}
Ramchoun, Hassan, Ghanou, Youssef, Ettaouil, Mohamed, Janati Idrissi, Mohammed Amine,
"Multilayer perceptron: Architecture optimization and training",
\textit{International Journal of Interactive Multimedia and Artificial Intelligence~…}, vol. , no. , pp. , 2016.

\bibitem{Hendrycks2016Gaussian}
Hendrycks, Dan, Gimpel, Kevin,
"Gaussian error linear units (gelus)",
\textit{arXiv preprint arXiv:1606.08415}, vol. , no. , pp. , 2016.

\bibitem{Lin2017Feature}
Lin, Tsung-Yi, Doll{\'a}r, Piotr, Girshick, Ross, He, Kaiming, Hariharan, Bharath, Belongie, Serge,
"Feature pyramid networks for object detection",
\textit{}, vol. , no. , pp. 2117--2125, 2017.

\bibitem{xie2021segformer}
Xie, Enze, Wang, Wenhai, Yu, Zhiding, Anandkumar, Anima, Alvarez, Jose M, Luo, Ping,
"SegFormer: Simple and efficient design for semantic segmentation with transformers",
\textit{Advances in neural information processing systems}, vol. 34, no. , pp. 12077--12090, 2021.

\bibitem{Howard2019Searching}
Howard, Andrew, Sandler, Mark, Chu, Grace, Chen, Liang-Chieh, Chen, Bo, Tan, Mingxing, Wang, Weijun, Zhu, Yukun, Pang, Ruoming, Vasudevan, Vijay, others,
"Searching for mobilenetv3",
\textit{}, vol. , no. , pp. 1314--1324, 2019.

\bibitem{he2016deep}
He, Kaiming, Zhang, Xiangyu, Ren, Shaoqing, Sun, Jian,
"Deep residual learning for image recognition",
\textit{}, vol. , no. , pp. 770--778, 2016.

\bibitem{ronneberger2015u}
Ronneberger, Olaf, Fischer, Philipp, Brox, Thomas,
"U-net: Convolutional networks for biomedical image segmentation",
\textit{}, vol. , no. , pp. 234--241, 2015.

\bibitem{wu2023samgeo}
Wu, Qiusheng, Osco, Lucas Prado,
"samgeo: A Python package for segmenting geospatial data with the Segment Anything Model (SAM)",
\textit{Journal of Open Source Software}, vol. 8, no. 89, pp. 5663, 2023.

\bibitem{Xu2023HiSup:}
Xu, Bowen, Xu, Jiakun, Xue, Nan, Xia, Gui-Song,
"HiSup: Accurate polygonal mapping of buildings in satellite imagery with hierarchical supervision",
\textit{ISPRS Journal of Photogrammetry and Remote Sensing}, vol. 198, no. , pp. 284--296, 2023.

\bibitem{Zhang2023Faster}
Zhang, Chaoning, Han, Dongshen, Qiao, Yu, Kim, Jung Uk, Bae, Sung-Ho, Lee, Seungkyu, Hong, Choong Seon,
"Faster segment anything: Towards lightweight sam for mobile applications",
\textit{arXiv preprint arXiv:2306.14289}, vol. , no. , pp. , 2023.

\end{thebibliography}

\section{Appendix}
\label{sec:appendix}
This section showcases part of the satellite imagery of testing data alongside predicted masks made by SAM-Adapter and FPN-MiT using Train-Small datasets without augmentation or upscaling. Some examples of errors in the predicted masks are indicated by red circles.

\begin{figure}[h]
    \centering
    \includegraphics[width=0.9\linewidth]{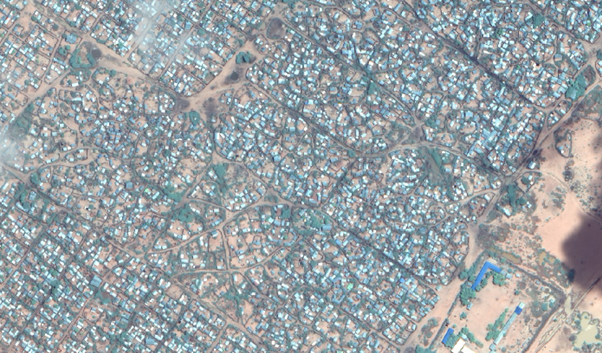}
    \caption{Satellite imagery of Dagahaley refugee camp.}
    \label{fig:figa1}
\end{figure}

\begin{figure}[h]
    \centering
    \includegraphics[width=0.9\linewidth]{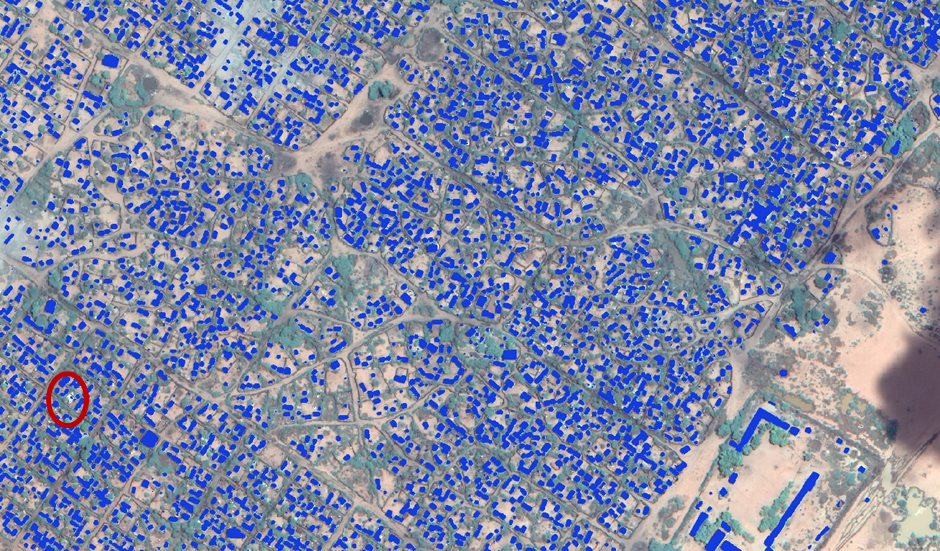}
    \caption{Predicted masks of SAM-Adapter trained on Train-Small dataset without upscaling in Dagahaley Refugee Camp.}
    \label{fig:figa2}
\end{figure}

\begin{figure}[h]
    \centering
    \includegraphics[width=0.9\linewidth]{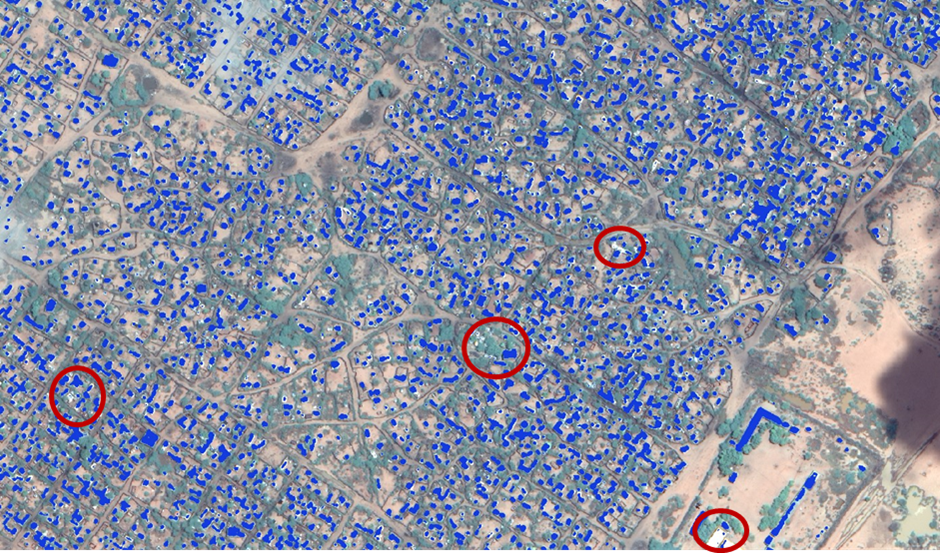}
    \caption{Predicted masks of FPN-MiT trained on Train-Small dataset without upscaling in Dagahaley Refugee Camp.}
    \label{fig:figa3}
\end{figure}

\end{document}